\documentclass[letterpaper, 10 pt, conference]{ieeeconf}

\IEEEoverridecommandlockouts
\usepackage{graphicx}
\usepackage{amsmath,amssymb} 
\usepackage{color}
\usepackage{bbm}
\usepackage{caption,setspace}
\usepackage{verbatim}
\usepackage{subfigure}
\usepackage{multirow}
\usepackage{array}
\usepackage{siunitx}
\usepackage{xcolor,colortbl}
\usepackage{rotating}
\usepackage{comment}
\usepackage{float}
\usepackage{comment}
\usepackage[utf8]{inputenc}
\usepackage{xspace}
\usepackage{hyperref}
\usepackage{url}
\newcommand{\merci}{n\nobreakdash-MeRCI\xspace}
\usepackage{soul}

\title{\LARGE \bf
\merci: A new Metric to Evaluate the Correlation Between Predictive Uncertainty and True Error
}

\author{
Michel Moukari$^{1,2}$, Loïc Simon$^{1}$, Sylvaine Picard$^{2}$ and Frédéric Jurie$^{1}$
\thanks{$^{1}$Michel Moukari, Loïc Simon and Frédéric Jurie are with Normandie Univ, UNICAEN, ENSICAEN, CNRS, 6 boulevard du Maréchal Juin 14000 Caen, France
        {\tt\small {\{michel.moukari, loic.simon, frederic.jurie\}}@unicaen.fr}}
\thanks{$^{2}$Michel Moukari and Sylvaine Picard are with Safran, 1 rue Geneviève Aube, 78114 Magny-les-Hameaux, France
        {\tt\small {\{michel.moukari, sylvaine.picard\}}@safrangroup.com}}
}

\usepackage{xspace}
\begin{document}

\def\ie{\emph{i.e.}\@\xspace}
\def\etc{\emph{etc.}\@\xspace}
\def\eg{\emph{e.g.}\@\xspace} \def\Eg{\emph{E.g.}\@\xspace}
\def\etal{\emph{et al.}\@\xspace}

\definecolor{num16}{RGB}{255,255,0}
\definecolor{num26}{RGB}{255,204,0}
\definecolor{num36}{RGB}{255,153,0}
\definecolor{num46}{RGB}{255,102,0}
\definecolor{num56}{RGB}{255,51,0}
\definecolor{num66}{RGB}{255,0,0}

\definecolor{num14}{RGB}{255,255,0}
\definecolor{num24}{RGB}{255,170,0}
\definecolor{num34}{RGB}{255,85,0}
\definecolor{num44}{RGB}{255,0,0}

\definecolor{mgray}{RGB}{170,170,170}

\maketitle
\thispagestyle{empty}
\pagestyle{empty}

\begin{abstract}
As deep learning applications are becoming more and more pervasive in robotics, the question of evaluating the reliability of inferences becomes a central question in the robotics community. This domain, known as {\em predictive uncertainty}, has come under the scrutiny of research groups developing Bayesian approaches adapted to deep learning such as Monte Carlo Dropout. Unfortunately, for the time being, the real goal of predictive uncertainty has been swept under the rug. Indeed, 
these approaches are solely evaluated in terms of raw performance of the  network prediction, while the quality of their estimated uncertainty  is not assessed. 
Evaluating such uncertainty prediction quality is especially important in robotics, as actions shall depend on the confidence in perceived information. 
In this context, the main contribution of this article is to propose a novel metric that is adapted to the evaluation of relative uncertainty assessment and directly applicable to regression with deep neural networks. To experimentally validate this metric, we evaluate it on a toy dataset and then apply it to the task of monocular depth estimation. 
\end{abstract}

\section{Introduction}
The outcome of deep learning these past few years -- capable of reaching and even sometimes surpass human performances, \eg for vision tasks~\cite{kokkinos2015surpassing,lu2015surpassing} --  offers novel avenues in robotics, but also raises some novel questions: as pointed out by \cite{SunderhaufBSHFL18}, {\em how much trust can we put in the predictions of a deep learning system?}, {\em how can we estimate the uncertainty in a deep network’s predictions?} are among the central questions that have to be addressed before deep learning can safely be used in robotics. 

While the robustness of deep learning algorithms to adversarial attacks has been very actively addressed in the recent literature (see \eg, \cite{Papernot2016limitations}), the question of predicting the uncertainty of predictions made by deep neural networks has received much less attention. Yet, for countless applications it is a matter of critical importance to know whether the estimate output of the algorithm is reliable or not, especially in robotics. In addition, the rare papers which addressed related questions do not evaluate the quality of the predicted uncertainty (\eg, \cite{liu2019NeuralRGBSensing} in the context of depth inference from RGB images).

In this paper we take the example of {\em depth estimation from single monocular images} \cite{liu2016LearningDepthSingle,liu2019NeuralRGBSensing}, which is useful for many tasks in robotics (\eg, obstacle avoidance, vehicle guidance, \etc).
For such a task, deep convolutional neural networks are trained to regress a depth map from an input RGB image.
If one wants to use such algorithms as  measurement tools in robotics applications, it is required to have, in addition to the predicted depth, an uncertainty map providing an estimate of the accuracy for each pixel.
Uncertainty prediction is indeed a very valuable information and also depends on the content of the images.
For example, it is likely that depth estimation on textureless regions will be less accurate, or, that dark images lead to less accurate estimations than well-exposed images.

In the literature of deep learning, some recent techniques have emerged to estimate uncertainties along with the network's predictions \cite{gal2016dropout,lakshminarayanan2017simple,kendall2017uncertainties}.
However, these predictive uncertainty techniques are evaluated either by their influence on the predictions performance, or qualitatively by visually checking the uncertainty maps. This is neither appropriate nor statistically accurate. 

 In the literature, the most common way to evaluate the predicted confidences is through a sparsification process (see \eg, \cite{xiaoyanhu2012QuantitativeEvaluationConfidence}), which is only measuring if the ranking of the confidences is correct. In domains such as forecasting, standard tools to evaluate predictive uncertainty \cite{uusitalo2015overview,gneiting2007probabilistic,gneiting2007strictly} do not allow to evaluate the quality of the uncertainty estimate independently of the predicted values.

In this context, the main contribution of this paper is  a novel evaluation metric that is well adapted to the evaluation of relative uncertainty assessment and is directly applicable to the current practice in deep learning. This metric is evaluated a on a toy dataset as well as in the context of depth estimation from a single monocular image.
This allows us to draw conclusions regarding several uncertainties methods including some of the most popular proposed in the literature and applied to this task.

\section{Related Work}

\paragraph{Uncertainty estimates}
 a common path to the estimate of uncertainties with deep networks is to transform  deterministic networks, in which each parameter is a scalar, into probabilistic Bayesian models, where parameters are replaced by probability distributions. Because the inference becomes intractable, several papers rely on approximate solutions. Among the most popular techniques, we count Probabilistic BackPropagation (PBP) \cite{hernandez2015probabilistic} or Monte Carlo Dropout \cite{gal2016dropout}. There are also some non-Bayesian ones such as Deep ensembles \cite{lakshminarayanan2017simple} or the learned attenuation loss proposed by~\cite{kendall2017uncertainties} to capture the aleatoric uncertainty in the data.

\paragraph{Assessment of predictive uncertainty}  in the literature, it is usually the quality of the prediction which is measured, not the quality of the uncertainty. Qualitative evaluations are commonly done visually with images of the estimated uncertainties. In terms of quantitative  evaluation, the most common metric is sparsification plots which measure how much the estimated uncertainty correlates with the true errors. If the data with the highest variance are removed gradually, the error should monotonically decrease \eg, \cite{xiaoyanhu2012QuantitativeEvaluationConfidence,GongY04}). As said before, one limitation of these metrics is that they  only evaluate the rankings of the confidences. 

The {\em Evaluating Predictive Uncertainty Challenge} \cite{quinonero2006evaluating} proposed two metrics adapted to regression tasks that are still commonly used. The normalized MSE aims at evaluating the quality of a prediction and uses the empirical variance of the test estimates to normalize the scores. This is therefore not a direct way to evaluate the predictive uncertainty. The second one, the average negative log predictive density (NLPD), directly takes into account the estimated uncertainty and penalizes both over and under-confident predictions.

A more in-depth treatment can be found in the forecasting literature.
 \cite{uusitalo2015overview} draw up an overview of  methods to estimate uncertainty with deterministic models such as: expert assessment, where an expert in the area rates the confidence of a method using his knowledge; model sensitivity analysis, determining how the output response of the model change w.r.t. the inputs and model's parameters; model emulation, consisting of approximating a complex model by a lower-order function, to then apply more easily a sensitivity analysis; spatiotemporal variability, using the spatiotemporal variability as an estimate of the variance; multiple models, where an ensemble of models are used to describe the same domain; and finally data-based approaches, where enough data are available to directly assess the statistical uncertainty related to the model outputs.

 Differently, \cite{gneiting2007probabilistic} propose to evaluate a predictive uncertainty forecast in terms of calibration and sharpness. Calibration refers to the statistical 
consistency between the predictive distributions and the observations; sharpness refers to the concentration of the predictive distributions.
Several graphical tools are proposed to visually evaluate if the probabilistic forecasts are calibrated and sharp. More recently, \cite{kuleshov2018accurate} propose a method to recalibrate regression algorithms so that the uncertainty estimates lie in the expected confidence interval. They also propose metrics to assess both calibration and sharpness of a predictive distribution. However, they consider them separately while we were driven by the will of assessing both these properties jointly. Following the tracks of \cite{gneiting2007strictly}, we are interested in the scoring rules functions that assess the quality of a forecast, maximizing sharpness subject to calibration.

\paragraph{Limitations of the previous methods for predictive uncertainty assessment} First, these methods do not allow to evaluate the uncertainty prediction independently of the inferred values; this is the whole distribution which is evaluated. Second, most of these metrics imply that the prediction is a probability distribution, which is seldom true in practice. Instead, existing deep learning methods usually predict uncertainty as being a scalar representing the variance of some predictions (\eg with Monte Carlo sampling). 
These remarks motivate the proposed metric allowing to evaluate prediction/uncertainty independently. We call our metric the  {\em normalized Mean Rescaled Confidence Interval} (\merci). It is based on the calculation of a scaling coefficient, for the interval centered on the prediction to encompass the ground-truth observation. It needs no prior on the distribution, allows to effectively rank different methods of estimations and is parametrized to be robust to outliers.

\section{Proposed Metric}

\subsection{Notations and definitions}
We assume a regression setting with a training set of $N$ samples denoted as ${\cal T}=\{(x_i,y_i^*)\}$ where $x_i \in \mathbb{R}^d$ is the input vector of dimension $d$ and $y_i^*\in \mathbb{R}$ the scalar value to be predicted.  

In a standard deterministic regression framework, the regressor is a function defined as $f_\theta: \mathbb{R}^d\rightarrow \mathbb{R}$, where $f_\theta(x)$ is an estimator of the true function $f(x)$, and  $\hat y = f_\theta(x)$ is the prediction made by the model. Learning such a regressor consists in  finding the parameters $\theta$ minimizing an empirical loss  defined by $\frac 1N \sum_{i=1}^N L(f_\theta(x_i), y_i^*)$ where $L(\cdot,\cdot)$ accounts for the proximity of its two arguments (e.g. a squared difference). 

In a probabilistic framework, we predict a whole distribution $P_y$ instead of a deterministic prediction.  This distribution has a probability density function $p_y(x)$ and a cumulative distribution $F_y$. 
In practice, $P_y$ is often characterized by its mean value, which is usually used as the prediction $\hat y$, and by its standard deviation $\sigma$. Different statistical approaches to perform such kinds of predictions in deep learning shall be reviewed in section~\ref{sec:methods}.

The previous notations assume the predictions to be defined through a distribution. However, most deep learning regression methods merely output a scalar value $\hat y$ along with a standard deviation $\sigma$. The proposed method is based on such  $\sigma$ values.

\subsection{Objective}
We denote the true error as the difference between estimated values and actual values: $\epsilon_i =  |\hat y_i-y_i^*|$. The Mean Absolute Error is defined as: $MAE = \frac{1}{N} \sum_{i=1}^{N}|\hat y_i-y_i^*|$.

Our aim is to define a metric  with the following properties:
\begin{itemize}
   \item Be scale independent and robust to outliers. Being scale independent is important as it avoids to advantage methods predicting large uncertainties. It normalizes uncertainties in terms of sharpness. 
   \item Reflect the correlation between the true error $\epsilon_i$  and the estimated uncertainty $\sigma_i$, independently of the estimated values $y_i$. 
    \item Give meaningful values. Being greater or equal to $1$ means the estimation $\sigma_i$ is worse than a constant uncertainty estimator (equivalent to have no estimation of the uncertainly). Being equal to $0$ means the prediction perfectly fits with the actual errors (lower bound).
\end{itemize}

\subsection{Mean Rescaled Confidence Interval} 
Regarding the first property, inspired by the notion of calibration and with scale invariance in mind, we propose to rescale all the $\sigma_i$'s by a common factor $\lambda$ so that the rescaled uncertainties $\tilde \sigma_i=\lambda \sigma_i$ are compatible with the observed errors, that is to say $|\hat y_i-y_i^*|\leq\tilde\sigma_i$. Robustness to outliers is achieved by relaxing this constraint such that it is verified by only a certain number of the samples. For example, one can choose to compute the MeRCI score on $95\%$ of the data considered to be reliable (inliners). In other words, we look for the smallest factor $\lambda^{\alpha}$ such that $\alpha\%$ of the observed errors are less than their corresponding rescaled predicted uncertainty.

Considering also the second property, the Mean Rescaled Confidence Interval that we propose is an average uncertainty, performed over the $\lambda\text{-rescaled}$ estimated uncertainties and averaged over the number of evaluated points $N$:
\begin{equation}
\textrm{MeRCI}^\alpha = \frac{1}{N} \sum_{i=1}^{N} \lambda^{\alpha} \, \sigma_{i}
\label{eqmerci}
\end{equation}

To obtain an efficient computation of the scaling factor, we first evaluate all ratios $\lambda_i=\frac{|\hat y_i-y_i^*|}{\sigma_i}$ and then extract the $\alpha^\text{th}$ percentile which corresponds exactly to the desired value.

MeRCI is inherently insensitive to scaling and allows us to compare methods with different ranges of uncertainties.

\subsection{Normalized Mean Rescaled Confidence Interval}

Regarding the third property, a range of relevant values can easily be exposed for this metric. On one hand, the MeRCI score is always lower bounded by the corresponding score for an ideal uncertainty prediction called the {\em oracle}.
This oracle merely predicts an uncertainty equal to the actual observed error, which yields to $\lambda^{\alpha}=1$ and a MeRCI score  reducing to the Mean Absolute Error (MAE).

On the other hand, any uncertainty prediction that performs worse than a  {\em constant} uncertainty predictor (i.e. a non-informative predictor) is clearly not worth pursuing. As a result, the score obtained by the constant predictor, $\max^{\alpha}(|\hat{y}_i-y_i|)$, provides a form of upper bound. $\max^{\alpha}$ is computed on the same extracted $\alpha^\text{th}$ percentile, as for MeRCI$^\alpha$. By defining \merci as:

\begin{equation}
 \text{\merci}^\alpha =  
    \frac{\frac{1}{N} \sum_{i=1}^{N} \, \lambda^{\alpha}\sigma_{i} - MAE}{\max^{\alpha}(|\hat{y}_i-y_i|) - MAE}
\end{equation}
we obtain the expected properties: it is scale invariant, is equal to 1 when the uncertainty predictor predicts a constant value for all data points, and is equal to 0 when the prediction is perfect (oracle). The score is therefore to be minimized. Moreover, when the uncertainty prediction is worse than using a constant uncertainty predictor, the score is greater than 1. This allows us to characterize the predictor as useless.

\merci is easy to interpret: a \merci value of 0.5 means the uncertainties are, on average, two times larger than the oracle. 

In its construction, the \merci score follows the track outlined by \cite{gneiting2007probabilistic}. It reflects the best possible sharpness under the constraint that the uncertainty estimate is calibrated, in the sense that it is consistent with $95\%$ of the samples. Being able to be robust to outliers is important as it avoids a few data points to strongly impact the whole measure, as shown in Section~\ref{sec:robust}. 
Also, the oracle is the sharpest possible uncertainty predictor once the estimated values $\hat y_i$ are fixed.

\section{Predictive uncertainty methods}
\label{sec:methods}

We consider several approaches to perform regression tasks with predictive uncertainty. First of all, we take the most used and popular methods in the literature of deep learning, among which we find Bayesian techniques.
Due to its small overhead, {\em Monte Carlo Dropout (\textbf{MCD})} \cite{gal2016dropout} is one such technique. At test time, the dropout is kept activated to perform Monte Carlo sampling. The final prediction is the average of the different estimates and  their standard deviation is used as a measure of uncertainty.

We also explore some popular ensemble techniques. In ensemble learning, multiple algorithms are used to obtain a predictive performance. Here, the different deterministic outputs are combined together (usually averaged) to obtain the final prediction and their standard deviation is associated to the degree of uncertainty of the prediction. A classical ensemble approach consists in training the same network several times with different initial random seeds. It is the approach employed in Deep Ensembles \cite{lakshminarayanan2017simple}, and we refer to this method as  {\em Multi Inits (\textbf{MI})}.
Another popular ensemble way to estimate uncertainty is  {\em \textbf{Bagging}} \cite{breiman1996bagging}. It involves using several subsets of the dataset to train several networks and compute the mean/variance of the different outputs.

In addition to these standard methods, inspired by state-of-the-art, we also propose several techniques to predict uncertainty in a deep learning framework. First, we propose to consider the network at different stages of the training as being as many different networks as there are epochs, once the plateau is reached. Assuming that, we can simulate an ensemble method coming with no additional cost, with standard training procedures. However, unlike snapshot ensembles \cite{huang2017snapshot}, we do not use a particular learning rate policy during the learning stage. Moreover, we aim to use this kind of ensemble technique seeking more to produce a meaningful uncertainty map than enhancing the overall prediction. We refer to this method as {\em Multi Epochs (\textbf{ME})}.
On a par with multiple initializations, we propose to use several architectures  and refer to this ensemble technique as  {\em Multi Networks (\textbf{MN})}. Here again, we average the outputs to obtain the prediction and use their standard deviation as the measure of uncertainty.
Finally, we think that because we want to measure the correlation between the uncertainty and the error, it is possible to learn a neural network to infer the confidence of a former neural network. The output of this network that learns the error will be therefore the uncertainty estimate paired with the prediction of the initial network. It is referred as  {\em Learned Error (\textbf{LE})}.

\section{Experimental validation on a toy dataset}
\label{sec:experiments_toy}

\subsection{Experimental protocol}

\begin{figure}[tb]
\centering
\captionsetup{justification=centering}
\includegraphics[width=0.95\columnwidth]{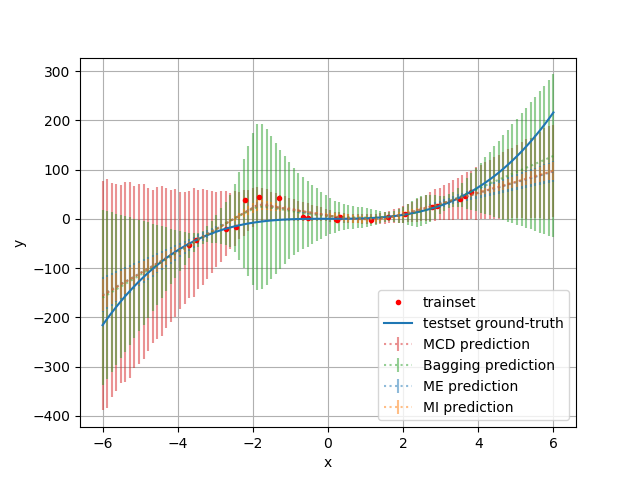}
\caption{The toy dataset.}
\label{toy_data}
\end{figure}

\begin{figure}[tb]
\centering
\captionsetup{justification=centering}
\includegraphics[width=0.95\columnwidth]{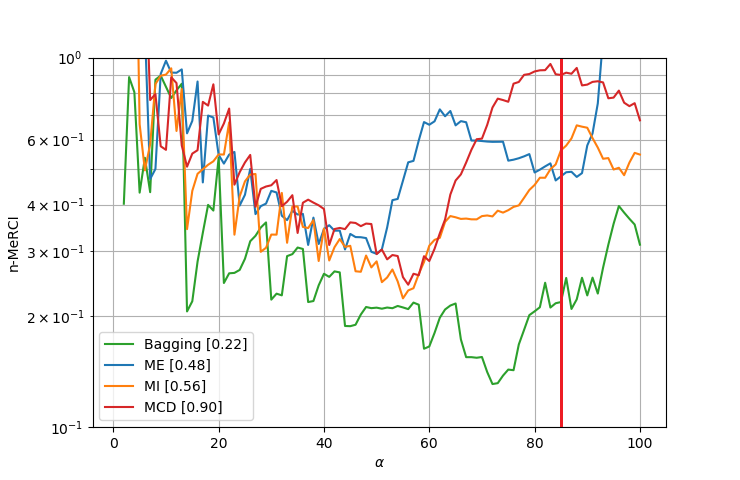}
\caption{\merci w.r.t. $\alpha$ parameter. Scores in brackets are displayed for $\alpha=85$ which corresponds to the actual inliners rate (marked by the red line).}
\label{n_merci_evolution}
\end{figure}

The toy dataset corresponds to a simple one-dimensional regression task. The data consists of 20 training examples, sampled uniformly in the interval $[-4;4]$. Their corresponding targets are generated as $y=x^3+\epsilon$, where $\epsilon \sim \mathcal{N}(0,3^2)$. We add a high bias to each training point in the interval $[-2.3;-1.3]$ to simulate some outliers, as illustrated in Figure \ref{toy_data}. A similar dataset has already been used to qualitatively evaluate predictive uncertainty in \cite{hernandez2015probabilistic} and \cite{lakshminarayanan2017simple}.

Given $x$, we learn to regress $y$ using a neural network with one hidden layer consisting of 100 neurons. We use the ReLU non-linearity and a dropout with dropping probability of 0.2 on the hidden layer.
We use 4 different techniques that we presented earlier: Multi Inits, Bagging, Monte Carlo Dropout and Multi Epochs. For each one we compute 20 runs. 
Their average predictions and uncertainty results are also illustrated in Fig. \ref{toy_data}. As expected, we see that the predictions are less accurate outside of the training set as well as in the outliers interval, with higher variance in these areas (displayed with $\times10$ scaling for better visualization).

\subsection{Robustness to outliers\label{sec:robust}}

Unlike classical metrics, \merci can be tailored towards obtaining more or less tolerance to erroneous data by tuning the percentile value.
To gain better understanding, we plot the evolution of \merci with respect to the chosen percentile in Fig. \ref{n_merci_evolution}. In this toy dataset, the ratio of corrupted data is known in advance and is equal to $15\%$ (3 data points).

As expected, the evolution of the \merci score decreases as the number of inliners increases, to some extent. Indeed, taking into account more correctly predicted points improves the score. However, although there is $85\%$ of outliers, we observe a deterioration around $\alpha=60\%$. This is due to the fact that most predictors present high variances around the outliers area, for correct estimations (see Fig. \ref{toy_data}). This encompasses around $40\%$ of the points.

In practice, the percentile should be chosen according to the number of supposed outliers in the dataset.
Therefore in later experiments we use the 95th percentile, giving some latitude to handle up to $5\%$ of possible outliers since we know that ground-truth is not perfect for the dataset we use.

\subsection{Comparisons of the uncertainty prediction methods on the toy dataset}

Finally, we evaluate the \merci scores obtained for each uncertainty estimation technique.
We expect Bagging to be the top technique because it benefits from external sources of data. 
We also expect Multi Epochs to be one of the worst. Indeed, the network is likely to converge to a local minimum and therefore there won't be much informative variance across epochs.

Although very noisy due to the number of data points, the plot confirms our intuitions. Indeed, Fig. \ref{n_merci_evolution} show the actual performances and looking at the specific value of $\alpha$=0.85, we see that: Bagging is best (n\nobreakdash-MeRCI=0.22), Multi Epochs is worse (n\nobreakdash-MeRCI=0.48) and the worst is Monte Carlo Dropout (n\nobreakdash-MeRCI=0.9). We also note that, on overall, predictions are far from being perfect. It is understandable as training points cover only a small part of the distribution, making estimation more difficult.
\section{Predictive uncertainties for  monocular depth estimation}

\subsection{Experimental protocol}

This section presents experiments on depth estimation from monocular images on the NYU-Depth v2 indoor dataset~\cite{Silberman:ECCV12}.
We use the set of 16K images proposed by \cite{moukari2018deep}.
Also, we use their hourglass-type architecture, which backbone is a dilated ResNet200 pre-trained on ImageNet for the encoder part, and 4 up-projection modules followed by a final convolution for the decoder.
We then add dropout with probability 0.2 after the first 3 up-projection modules, in order to perform Monte-Carlo sampling at test time for the corresponding experiment.

\subsection{Qualitative evaluation}

\begin{figure*}[tb]
\centering
\subfigure[input]{\includegraphics[width=0.25\textwidth]{img/input_403.png}\label{common_prediction_a}}
\subfigure[prediction]{\includegraphics[width=0.25\textwidth]{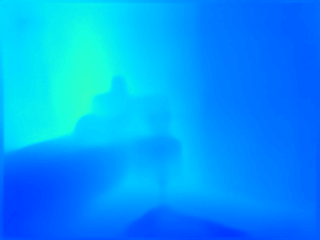}\label{common_prediction_c}}
\subfigure[absolute error]{\includegraphics[width=0.25\textwidth]{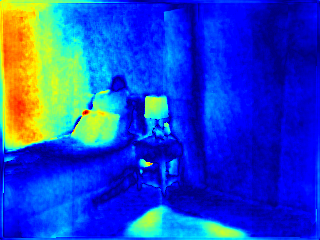}\label{common_prediction_d}}\\

\subfigure[Multi Epochs \newline $\text{\merci}^{95}=0.45$, $\text{MAE}^{95}=0.13$]{\includegraphics[width=0.25\textwidth]{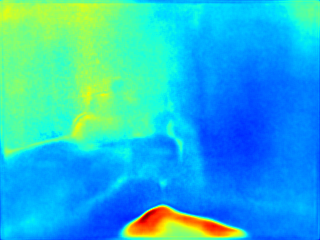}\label{std_ME}}
\subfigure[Monte Carlo Dropout \newline $\text{\merci}^{95}=0.47$, $\text{MAE}^{95}=0.14$]{\includegraphics[width=0.25\textwidth]{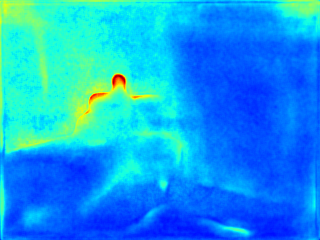}\label{std_MCD}}
\subfigure[Multi Inits \newline $\text{\merci}^{95}=0.60$, $\text{MAE}^{95}=0.12$]{\includegraphics[width=0.25\textwidth]{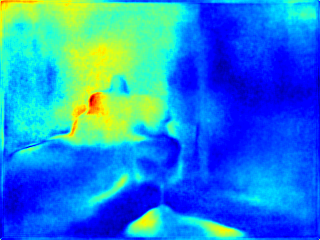}\label{std_MI}}\\

\subfigure[Multi Networks \newline $\text{\merci}^{95}=0.63$, $\text{MAE}^{95}=0.21$]{\includegraphics[width=0.25\textwidth]{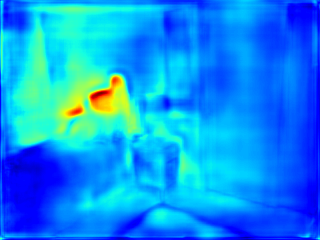}\label{std_MN}}

\subfigure[Bagging \newline $\text{\merci}^{95}=0.95$, $\text{MAE}^{95}=0.12$]{\includegraphics[width=0.25\textwidth]{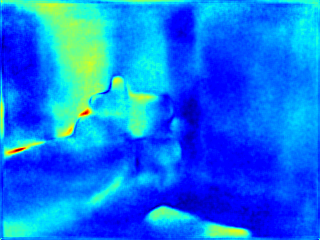}\label{std_KF}}
\subfigure[Learned Error \newline $\text{\merci}^{95}=4.2$, $\text{MAE}^{95}=0.14$]{\includegraphics[width=0.25\textwidth]{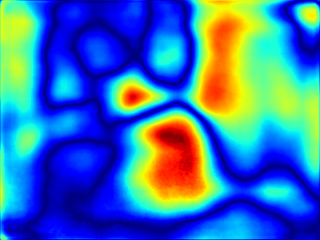}\label{std_LE}}\\

\caption{A NYU Depth v2 test case. First row: the test image along with a predicted depth map and the corresponding absolute errors. Remaining rows:  predictive uncertainties, ranked according to their \merci scores.}
\label{different_uncertainties}
\end{figure*}

We first show a qualitative evaluation obtained for the monocular depth estimation. In Fig.~\ref{different_uncertainties}, we show an example of a test image \ref{common_prediction_a}, the prediction of one predictor \ref{common_prediction_c} and finally the absolute error between this prediction and the ground truth \ref{common_prediction_d}; all methods predictions are close to each other so we choose to only display one for visualization. According to the absolute error map, a few demarcated regions should be marked as uncertain: the carpet, the bedside lamp, the pillow and the window.

\begin{figure}[tb]
\centering
\captionsetup{justification=centering}
\includegraphics[width=0.95\columnwidth]{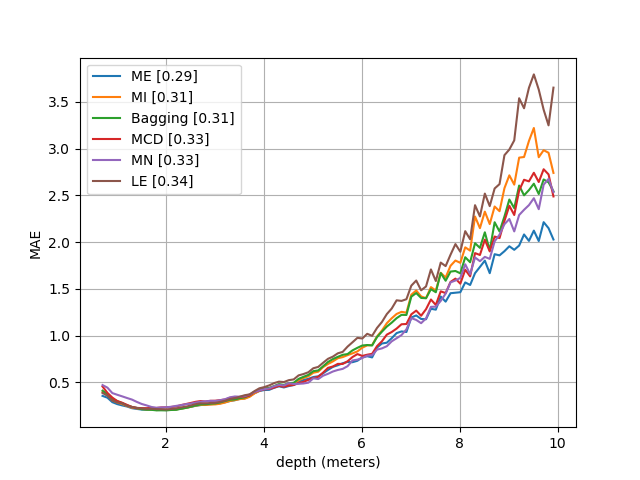}
\caption{MAE w.r.t. depth intervals for 95\% of the points. Average MAE scores are displayed in brackets.}
\label{mae_fdis_95}
\end{figure}

\begin{figure}[tb]
\centering
\captionsetup{justification=centering}
\includegraphics[width=0.95\columnwidth]{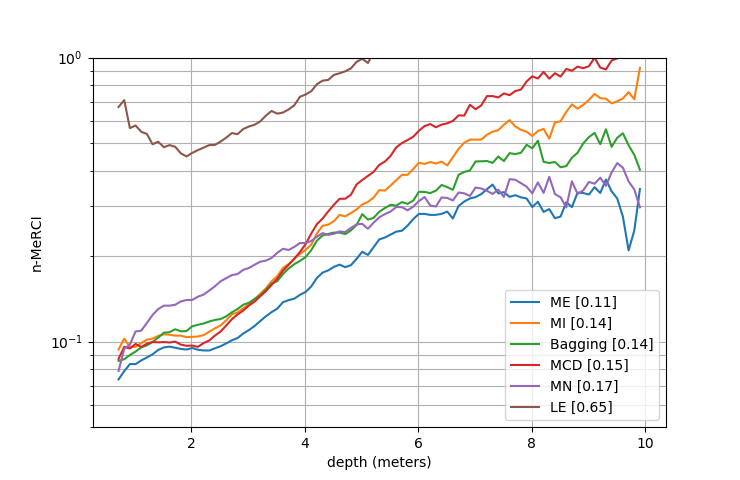}
\caption{\merci w.r.t. depth intervals ($\alpha$=95\%). \\ Average $\text{\merci}^{95}$ scores are displayed in brackets.}
\label{merci_fdis_95}
\end{figure}

We display the 6 uncertainty estimates obtained with the different methods in Fig.~\ref{different_uncertainties}.
Only two methods show high uncertainty in the region corresponding to the window: namely Multi Epochs and Monte Carlo Dropout.
This region is clearly the most important as it is the one with the highest errors according to the error map Fig.~\ref{common_prediction_d}.
Furthermore, Multi Epochs also displays high uncertainty at the carpet location, while Monte Carlo Dropout misses this place.
This is the reason why \merci ranked these methods as the two top ones, but favors Multi Epochs.

Meanwhile, we see that Multi Networks, Multi Inits or Bagging have poor correlation with the true error. Indeed, they present high uncertainty around the pillow but misses the other important spots: either the carpet and bedside lamp for Multi Networks or the window for Multi Inits and Bagging.
Multi Networks also captures variance on the edges of the RGB image which is inconsistent with the true error.
This is due to the fact that some of the different networks involved in the estimations produce sharper outputs but at the price of some artifacts transferred from the RGB images \cite{moukari2018deep}.

Eventually, Learned Error completely fails to capture the relevant error information, which is correctly evaluated by \merci.

\subsection{Quantitative evaluation\label{sec:depth-qual}}
\label{section:quantitative}

Quantitative results are presented in Fig. \ref{mae_fdis_95}~(MAE) and Fig. \ref{merci_fdis_95}~(\merci). We withdraw $5\%$ of the worst points in these experiments and \merci is computed locally for depth intervals of size 0.1 meter. First of all, it makes sense to inspect the MAE, shown on Fig.~\ref{mae_fdis_95}, as selecting a method purely on its ability to predict the uncertainty, without taking care of the precision, would be risky. The different methods are very close in terms of precision, with a slight advantage to Multi Epochs (MAE=0.29). The worse one is Learned Error, which makes sense as it does not average several predictions (MAE=0.34). We also observe that MAE increase as depth grows. Fig. \ref{merci_fdis_95} shows \merci scores for the different methods. In the same way as before, we see that \merci worsen as depth increase, which means that uncertainties are less accurate for higher depth intervals. This makes sense as these intervals are statistically less represented in the NYU dataset.
Similarly to the single-image qualitative evaluation, \merci underlines Learned Error as the worst forecast (\merci=0.65) and Multi Epochs as the best one (\merci=0.11).
This favored rank is supported by the fact that training a large network on a complex task and with a stochastic minimization scheme naturally yields oscillations around a minimum.
What is more, along the last epochs, the network will focus on learning what is still improvable: \ie regions with high errors.
Thus, using the standard deviation of estimations obtained across epochs allows to capture these variations and highlights error-prone regions.
On the other hand, the results obtained with LE show how difficult it is to regress uncertainties from images with a feed-forward network, without using any statistical properties of the main network.

\section{Conclusions}
This paper has proposed a  metric, the Normalized Mean Rescaled Confidence Interval (\merci), dedicated to the assessment of predictive uncertainty. We  showed that the proposed metric behaves as expected on a toy dataset, while giving latitude to be robust to outliers if needed (thanks to its percentile parameter).  \merci has the advantage of being directly applicable to regression tasks with deep learning, without needing any prior on the distribution. Finally, we used \merci to evaluate the merits of various uncertainty estimation methods applied to the problem of monocular depth estimation, and ranked them w.r.t. such merits. We showed that Multi Epochs, which is one of the fastest, is the one giving the best prediction of the uncertainties.  

\bibliography{egbib}
\bibliographystyle{IEEEtran}

\end{document}